\documentclass{article}

\usepackage[dblblindworkshop,final]{neurips_2026}
\usepackage[utf8]{inputenc}
\usepackage[T1]{fontenc}
\usepackage{hyperref}
\usepackage{url}
\usepackage{booktabs}
\usepackage{amsfonts}
\usepackage{amsmath}
\usepackage{nicefrac}
\usepackage{microtype}
\usepackage{xcolor}
\usepackage{graphicx}
\usepackage{adjustbox}
\usepackage{float}
\usepackage[hypcap=false]{caption}
\usepackage{pgfplots}
\usepgfplotslibrary{fillbetween}
\pgfplotsset{compat=1.18}

\title{
The Coverage Selection Gap: Ranking, Not Generation,
Limits LLM Rubric Construction
}

\workshoptitle{Long-Context Foundation Models (LCFM)}

\author{%
  {\normalfont\bfseries\normalsize
  Ali Karkehabadi$^{1,3}$ \hspace{0.6em}
  Jamshid Hassanpour$^{2}$ \hspace{0.6em}
  Rafael E. Banchs$^{3,4}$}\\[5pt]
  {\normalfont\small
  $^{1}$University of California, Davis \quad
  $^{2}$Georgia Institute of Technology}\\[2pt]
  {\normalfont\small
  $^{3}$Intapp \quad
  $^{4}$San Jose State University}\\[5pt]
  {\normalfont\small
  $^{1}$\text{akarkehabadi@ucdavis.edu} \quad
  $^{2}$\text{jamshid@gatech.edu}}\\[2pt]
  {\normalfont\small
  $^{3}$\text{\{ali.karkehabadi,rafael.banchs\}@intapp.com}}\\[2pt]
  {\normalfont\small
  $^{4}$\text{rafael.banchs@sjsu.edu}}%
}

\begin{document}

\maketitle

\begin{abstract}

Task-specific rubrics are increasingly used to evaluate
open-ended language-model outputs, yet constructing
comprehensive evaluation criteria from complex source
documents remains challenging.
We study automatic rubric construction in professional
mergers and acquisitions (M\&A) due diligence, where
evaluation criteria must be inferred from task instructions
and supporting documents without access to an
expert-authored reference rubric.
We introduce a multi-axis generation framework that
explores complementary structural and free-form views
of the source evidence to construct a broad pool of
candidate evaluation criteria.
To distinguish failures of criterion discovery from
failures of candidate selection, we introduce the
\emph{coverage--selection gap}, which measures the
difference between reference-free ranking and a
reference-informed greedy diagnostic under a fixed
selection budget.
Developed on 16 tasks and evaluated on four held-out
tasks, our framework achieves 0.876 macro recall
against senior-authored reference rubrics under an
LLM-based criterion-equivalence protocol.
Ablation studies further show that introducing new
structural search axes improves recall by 6--10
percentage points, whereas increasing generation
volume along existing search directions yields
at most 1.5 points.
These findings suggest that, once a high-coverage
candidate pool has been constructed, effective
reference-free selection becomes a substantial
bottleneck in document-grounded rubric construction.

\end{abstract}

\section{Introduction}

Large language models (LLMs) are increasingly used to evaluate open-ended
outputs for which exact-match and single-reference metrics are inadequate.
Rubric-based judges can represent task-specific notions of quality, but they
generally assume that the evaluation criteria are already available
\cite{kim2024prometheus,ye2025conj,xiong2026multicrit}. The quality
of evaluation therefore depends not only on how faithfully a judge applies a
rubric, but also on whether the rubric captures the requirements that matter. This specification problem is particularly challenging in document-intensive
professional workflows such as M\&A due diligence, where evaluation criteria
depend on the task and its supporting documents. A reviewer may need to verify
contractual obligations, reconcile entities across documents, identify missing
financial information, or determine whether conclusions are supported by the
evidence. Such requirements cannot be recovered from a fixed checklist, while
manually constructing task-specific rubrics is expensive. We therefore study
\emph{document-grounded automatic rubric construction}: given a task instruction
$t$ and document collection $D=\{d_1,\ldots,d_k\}$, generate evaluation
criteria $C=\{c_1,\ldots,c_n\}$ without access to the expert-authored reference
rubric $R^\star$. Unlike prior LLM-as-a-Judge work, which assumes the evaluation criteria are
known, we ask whether those criteria can be recovered directly from the task
and its supporting evidence. Existing work focuses on applying a given rubric,
despite known judge biases and inconsistencies
\cite{yang2025rbd,lai2026biasscope};
we instead address the upstream problem of discovering the rubric itself. Our framework combines five complementary generation processes that examine
different structural views of the task and source documents, producing a shared
candidate pool that is ranked without access to the reference rubric. To distinguish these
cases, we compare a deployable reference-free ranking with a greedy
reference-informed ordering of the \emph{same} candidate pool. The difference
between them, which we call the \emph{coverage--selection gap}, measures the
coverage recoverable through improved selection alone. 

On four held-out
professional due-diligence tasks, our candidate pools achieve 0.876 macro
recall against senior-authored rubrics. Yet at $k=60$, the reference-free
ranker recovers only 0.349 recall, while the reference-informed ordering
reaches the full 0.876 pool ceiling, revealing a 52.7 percentage-point
coverage--selection gap. Ablation studies further show that diversifying
document search contributes substantially more coverage than increasing
generation volume, and that the resulting candidate pools complement strong
single-call generation.

\paragraph{Research Questions.}

We organize our investigation around three research
questions.

{RQ1: Candidate discovery.}
To what extent can complementary document-search
strategies improve the coverage of evaluation
requirements compared with single-call generation?

{RQ2: Candidate selection.}
Given a high-coverage candidate pool, how effectively
can reference-free ranking recover relevant criteria
under a limited selection budget?

{RQ3: Generation versus selection.}
How much of the missing coverage in a selected rubric
is attributable to criteria that were never generated,
versus criteria that were generated but ranked below
the selection budget?

Our contributions are threefold:
(1) we formulate document-grounded rubric construction as an upstream
evaluation problem and develop a multi-axis framework for high-coverage
candidate discovery;
(2) we introduce the \emph{coverage--selection gap}, which distinguishes
criteria that were never generated from those generated but not selected under
a practical budget; and
(3) we show that structural search diversity contributes substantially more
coverage than additional generation volume while remaining complementary to
strong single-call generation.

\vspace{3 pt}

\section{Related Work}

\paragraph{LLM-based evaluation and criterion-conditioned judging.}
LLMs are increasingly used to evaluate outputs that fixed metrics capture poorly.
Prometheus evaluates responses under customized rubrics
\cite{kim2024prometheus}, while Auto-J and Con-J learn natural-language
judgments from demonstrations or preferences
\cite{li2024autoj,ye2025conj}. Other work improves judge reasoning through
reinforcement learning, explicit deliberation, and process-level or multimodal
evaluation
\cite{whitehouse2026j1,guo2025rrm,
chae2025webshepherd,xiong2025llavacritic,
zhang2025upme}. Multi-Crit studies whether judges can follow
multiple criteria simultaneously \cite{xiong2026multicrit}. These methods
assume an available evaluation specification; we instead recover a
task-specific specification from source evidence.

\paragraph{Reliability of learned judges.}
LLM judges exhibit positional and preference biases
\cite{yang2025rbd}, can diverge from human assessments
\cite{polo2025bridge}, and remain sensitive to semantic and stylistic variation
\cite{wei2025semantickg}. Reward-model studies likewise show
that held-out accuracy need not predict downstream policy quality
\cite{wen2025rethinkingrm,frick2025ppe}. We address a different source of
evaluation failure: an evaluator cannot apply a criterion absent from its
rubric.

\paragraph{Professional legal-agent benchmarks.}
Harvey LAB provides more than 1,200 realistic legal-work tasks, each pairing an
instruction, source-document data room, and hand-authored evaluation rubric
\cite{harveylab2026}. We use a document-grounded subset to study rubric
construction.

\paragraph{From criterion application to rubric construction.}
Existing rubric-conditioned, generative, and process-level evaluators
obtain their specifications from rubrics, annotations, preferences,
or supervision
\cite{kim2024prometheus,xiong2026multicrit,ye2025conj,
chae2025webshepherd}. We instead infer evaluation criteria from task
instructions and source documents, then select a usable subset from
a larger candidate pool. This distinguishes requirements that were
never generated from those generated but not selected under a
practical budget.

\paragraph{Automatic rubric construction and selection.}
Recent work also automates rubric construction. OpenRubrics generates
criteria from preferred and rejected responses and filters them for
preference consistency \cite{liu2025openrubrics}.
Auto-Rubric uses verification-driven refinement and compresses
criteria through an information-theoretic objective
\cite{xie2025autorubric}. GenRubric uses reinforcement learning
and cross-rubric comprehensiveness signals to improve generation
\cite{chen2026genrubric}. Unlike these approaches, we explicitly
measure the gap between available candidate coverage and the
coverage recovered by reference-free selection, using a
reference-informed greedy diagnostic on fixed candidate pools.

\section{Method}
\label{sec:method}

\subsection{Problem Setup and Framework Overview}
\label{sec:setup}

\begin{figure*}[b]
    \centering
    \IfFileExists{figures/Method_Ha.png}{\includegraphics[width=\linewidth]{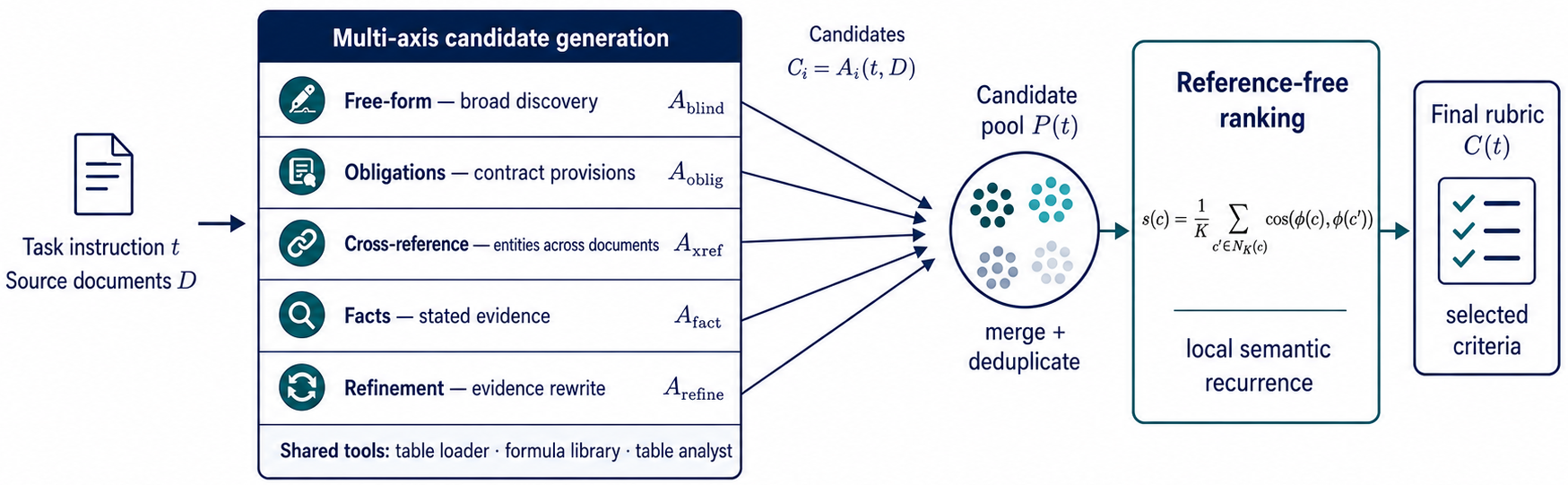}}{\fbox{\parbox{0.88\linewidth}{\centering Figure asset pending transfer: Method\_Ha.png}}}
    \caption{Overview of the rubric-construction pipeline. Five complementary generation processes examine the task instruction and source documents, populate a shared candidate pool, and rank the resulting criteria without access to the reference rubric before producing the final rubric.}
    \label{fig:pipeline}
\end{figure*}

Figure~\ref{fig:pipeline} shows the overview of the method. Given a task instruction $t$ and a document collection
$D=\{d_1,\ldots,d_m\}$, our goal is to construct a task-specific rubric
\[
C=\{c_1,\ldots,c_n\},
\]
whose criteria capture the requirements that should be used to evaluate the
completed deliverable. The expert-authored reference rubric
$R^\star=\{r_1,\ldots,r_q\}$ is never observed during rubric construction and
is used only for evaluation. A generated criterion $c$ is considered to cover a reference criterion $r$
when the two express the same underlying requirement, independent of wording
or surface form. The task therefore has two distinct components: discovering
relevant criteria from $(t,D)$ and selecting a usable subset from the resulting
candidate set. The final system uses five complementary generation processes that examine different structural views of the task instruction and source documents, producing a broad candidate pool ranked without access to $R^\star$.

Let $\mathcal{A}=\{A_1,\ldots,A_M\}$ denote the five processes in the final system. Each produces
{\footnotesize
\[
C_i=A_i(t,D),
\]
}
and their outputs form the candidate pool
\begingroup\footnotesize
\begin{equation}
\label{eq:pool_construction}
P(t)=
\operatorname{Dedup}
\left(
\bigcup_{i=1}^{M} C_i
\right).
\end{equation}
\endgroup
The candidate pool in Eq.~\eqref{eq:pool_construction} is ranked using a reference-free score
$s(c)$ and returns the highest-ranked criteria under the desired output
budget.

\subsection{Multi-Axis Candidate Generation}
\label{sec:generation}

\paragraph{Development-stage requirement abstraction.}
To represent recurring due-diligence requirements without retaining deal-specific
answers, development rubrics were organized into 48 broad review archetypes and
2,263 finer-grained requirement atoms. Atomization removes transaction-specific
literals while retaining the operation to be checked (e.g., a figure, provision,
assessment, benchmark, or computation) and the entity to which it applies.
The shared library of 119 computation templates comprises the computation-tagged
atoms; it is not an additional generation process.

\paragraph{Development resources and data separation.}
The requirement abstractions and computation templates
were developed using the 16 development tasks, which also informed
the generation procedures and fixed methodological choices.
The four held-out tasks were excluded from development.
At evaluation time, the system receives only the held-out
task instruction and source documents; the expert-authored reference
rubric remains hidden from generation and reference-free selection
and is used solely for post-hoc evaluation and reference-informed
diagnostic analysis. This separation is intended to prevent direct
reference-rubric leakage into the evaluated procedures.

\paragraph{Illustrative abstraction.}
For example, a historical training criterion asks whether a related-party lease
rate exceeds a comparable market rate and by how much annually. The reusable
operations are to extract the contractual rate, identify the documented
benchmark, assess materiality, and calculate annual excess rent when the
required area and rate figures are available. The deal-specific figures are
not stored as the answer to a future task. We diversify candidate discovery by changing the \emph{unit of search}.
All five processes operate on the task instruction and source documents
without access to the reference rubric, but search different structural views
of the evidence. The free-form process $A_{\mathrm{blind}}$ directly proposes criteria from the
task and documents; we union four independent generation seeds to broaden its
coverage. $A_{\mathrm{oblig}}$ iterates over relevant contractual
provisions, $A_{\mathrm{xref}}$ follows entities across schedules and
documents, and $A_{\mathrm{fact}}$ starts from explicit facts and tests whether
they are appropriately reflected in the target analysis. The final process, $A_{\mathrm{refine}}$, operates on an existing candidate
together with its supporting evidence to recover specificity lost during
initial generation, including figures, named provisions, or other
evidentiary details.
Table~\ref{tab:agents} summarizes the search unit and role of each generation process.

% \paragraph{Illustrative abstraction.}
% For example, a historical training criterion asks whether a related-party lease
% rate exceeds a comparable market rate and by how much annually. The reusable
% operations are to extract the contractual rate, identify the documented
% benchmark, assess materiality, and calculate annual excess rent when the
% required area and rate figures are available. The deal-specific figures are
% not stored as the answer to a future task. Rather than leaving the choice of what to inspect entirely to the model,
% we diversify candidate discovery by changing the \emph{unit of search}.
% All five processes operate on the task instruction and source documents
% without access to the reference rubric, but search different structural views
% of the evidence. The free-form process $A_{\mathrm{blind}}$ directly proposes criteria from the task and documents; we union four independent generation seeds to broaden its
% coverage. The three structural processes instead enumerate bounded units in
% the evidence. $A_{\mathrm{oblig}}$ iterates over relevant contractual
% provisions, $A_{\mathrm{xref}}$ follows entities across schedules and
% documents, and $A_{\mathrm{fact}}$ starts from explicit facts and tests whether
% they are appropriately reflected in the target analysis. The final process, $A_{\mathrm{refine}}$, operates on an existing candidate
% together with its supporting evidence to recover specificity lost during
% initial generation, including figures, named provisions, or other
% evidentiary details.

\begin{table}[h]
\centering
\caption{Generation processes differ in their search
units rather than only in prompt wording.}
\label{tab:agents}

\small
\setlength{\tabcolsep}{3pt}
\renewcommand{\arraystretch}{1.15}

\begin{tabular}{@{}lll@{}}
\toprule
\textbf{Process} & \textbf{Search unit} & \textbf{Role} \\
\midrule
$A_{\mathrm{blind}}$  & Unrestricted & Free-form discovery \\
$A_{\mathrm{oblig}}$  & Contract provision & Obligation checks \\
$A_{\mathrm{xref}}$   & Cross-doc. entity & Consistency checks \\
$A_{\mathrm{fact}}$   & Stated fact & Disclosure coverage \\
$A_{\mathrm{refine}}$ & Candidate + evidence & Specificity recovery \\
\bottomrule
\end{tabular}

\end{table}

% The free-form process $A_{\mathrm{blind}}$ directly proposes criteria from the
% task and documents; we union four independent generation seeds to broaden its
% coverage. The three structural processes instead enumerate bounded units in
% the evidence. $A_{\mathrm{oblig}}$ iterates over relevant contractual
% provisions, $A_{\mathrm{xref}}$ follows entities across schedules and
% documents, and $A_{\mathrm{fact}}$ starts from explicit facts and tests whether
% they are appropriately reflected in the target analysis.

% The final process, $A_{\mathrm{refine}}$, operates on an existing candidate
% together with its supporting evidence. Its purpose is to recover specificity
% that may be lost during initial generation, including figures, named
% provisions, or other evidentiary details.

\paragraph{Evidence-grounded specificity restoration.}
Unlike the discovery processes, $A_{\mathrm{refine}}$ revisits an existing
candidate and its supporting evidence to restore omitted amounts, dates, or
provision references. In a development-task diagnostic on
\texttt{shadow\_08}, the reported number of figures per claim increased
from 0.90 to 3.83 after refinement. This describes specificity of the
generated text, not independently verified criterion correctness.

More explicitly, the pool in Eq.~\eqref{eq:pool_construction} is
\begingroup\footnotesize
\begin{equation}
\label{eq:pool_expanded}
\adjustbox{max width=0.78\columnwidth}{$\displaystyle P(t)=\operatorname{Dedup}\!\bigl(C_{\mathrm{blind}}\cup C_{\mathrm{oblig}}\cup C_{\mathrm{xref}}\cup C_{\mathrm{fact}}\cup C_{\mathrm{refine}}\bigr).$}
\end{equation}
\endgroup
Each candidate retains its generating-process label, which we use for
subsequent attribution and ablation analyses.

\paragraph{Shared document tools.}
All generation processes use the same document-processing layer. A
table-aware loader preserves structured rows and headers that would otherwise
be absent from plain document text. A fixed library of 119 computation
templates supports recurring quantitative checks when the required inputs are
present. Finally, a table-analysis tool reconciles row-level entities and
quantities against totals or statements in surrounding narrative. These tools
expose evidence to the generation processes but do not themselves determine
which rubric criteria should be produced.

\paragraph{Cross-source table reconciliation.}
The table-analysis tool enables criteria that check agreement between
narrative statements and tabular evidence rather than merely restating
individual figures. For example, a narrative reporting seven retention
awards totaling \$1.85M and a supporting schedule listing six awards
totaling \$1.775M motivates a criterion requiring the reviewer to flag
and resolve the discrepancy. This is an illustrative methodological example,
not a newly evaluated held-out prediction.

\subsection{Reference-Free Candidate Selection}
\label{sec:selection}

The candidate pool is deliberately optimized for coverage and can therefore be
substantially larger than a practical review rubric. We rank candidates using
a local redundancy statistic that requires neither the reference rubric nor
additional LLM judgments.

Let $\phi(\cdot)$ be a frozen sentence-embedding function and
$\mathcal{N}_K(c)$ the $K$ nearest neighbors of candidate $c$ within
$P(t)\setminus\{c\}$. We define
\begingroup\footnotesize
\begin{equation}
\label{eq:recurrence_score}
s(c)
=
\frac{1}{K}
\sum_{c'\in\mathcal{N}_K(c)}
\cos\!\left(
\phi(c),
\phi(c')
\right),
\end{equation}
\endgroup
with $K=5$ (Eq.~\eqref{eq:recurrence_score}).

The score measures repeated semantic expression in the pool, not cross-process convergence. Deduplication retains only the first-seen origin, so provenance cannot distinguish recurrence across processes from recurrence across seeds. We therefore interpret $s(c)$ as a pool-level recurrence signal and rank candidates in decreasing order.
\paragraph{Interpretation of the ranking score.}
The recurrence score measures the extent to which
a candidate is semantically supported by neighboring
candidates in the generated pool.
Importantly, semantic recurrence should not be
interpreted as a direct measure of criterion
importance or marginal coverage.
A frequently generated requirement may receive
a high score even when it overlaps substantially
with other selected criteria.
Conversely, a rare but important requirement may
receive a low score despite contributing distinct
evaluation coverage.
We therefore treat local semantic recurrence as
a reference-free selection heuristic rather than
an explicit estimator of the additional reference
coverage contributed by each candidate.
This distinction motivates our subsequent analysis
of the coverage--selection gap.
For the filtered output, we retain candidates above a fixed score quantile,
\begingroup\footnotesize
\begin{equation}
\label{eq:filtered_candidates}
C_{\alpha}(t)
=
\left\{
c\in P(t):
s(c)\ge q_{\alpha}
\right\},
\end{equation}
\endgroup
where $q_{\alpha}$ is the $\alpha$-quantile and $\alpha=30\%$; equivalently, we discard the bottom 30\%. Both $K$ and $\alpha$ are fixed on development data. The full ordering is used for Recall@$k$ and the filtered set for set-level metrics.

The selection stage only reorders or removes criteria already present in the
pool; it never generates additional rubric content. This distinction allows
generation quality and selection quality to be analyzed separately.\\\\\\

\section{Evaluation Protocol}
\label{sec:evaluation}
\label{sec:equivalence}
\label{sec:cs_gap}

We evaluate criterion-level semantic equivalence rather than lexical overlap. For each reference criterion $r\in R^\star$, embeddings retrieve the $K_J=6$ nearest candidates for shortlisting only; an LLM judge decides whether $c$ expresses the same requirement as $r$, denoted $c\sim r$. Recall is
\begingroup\footnotesize
\begin{equation}
\label{eq:recall}
\adjustbox{max width=0.78\columnwidth}{$\displaystyle \operatorname{Recall}(C,R^\star)=\frac{|\{r\in R^\star:\exists c\in C,\ c\sim r\}|}{|R^\star|}.$}
\end{equation}
\endgroup
Precision is the fraction of candidates matching at least one reference criterion, using a separately constructed direction-specific reverse shortlist. Metrics are conditional on an LLM judge using the same deployment as the generators; no independent human validation was conducted.

Recall@$k$ is the fraction of references covered by the first $k$ candidates. We compare the deployable ranking with a non-deployable greedy reference-informed diagnostic that repeatedly selects the largest marginal reference coverage:
\begingroup\footnotesize
\begin{equation}
\label{eq:coverage_selection_gap}
\adjustbox{max width=0.78\columnwidth}{$\displaystyle G(k)=\operatorname{Recall@}k(\pi_{\mathrm{greedy}})-\operatorname{Recall@}k(\pi_{\mathrm{ref\text{-}free}}).$}
\end{equation}
\endgroup
The gap in Eq.~\eqref{eq:coverage_selection_gap} indicates existing coverage not surfaced within budget. We next
give the complete matching and diagnostic procedures used to compute this
quantity.

\subsection{Criterion-Equivalence Protocol}
\label{sec:equivalence_protocol}

Exhaustively judging every candidate--reference pair would be expensive and
would devote most calls to obviously unrelated pairs. We therefore construct
direction-specific shortlists with sentence embeddings. For recall, each
reference criterion retrieves its $K_J=6$ nearest candidates, which are
evaluated under a recall-oriented prompt. Embeddings determine only the
shortlist; the LLM judge makes the semantic-equivalence decision. For
precision, each candidate independently retrieves its six nearest reference
criteria and is evaluated under a directionally inverted usefulness prompt.
This asymmetry matters because the question ``does any candidate cover this
reference requirement?'' is not operationally identical to ``does this
candidate correspond to any reference requirement?''

Let $J(c,r)=1$ when the judge considers candidate $c$ and reference criterion
$r$ equivalent. Precision is then
\begingroup\footnotesize
\begin{equation}
\label{eq:precision}
\adjustbox{max width=0.78\columnwidth}{$\displaystyle \operatorname{Precision}(C,R^\star)=\frac{1}{|C|}\sum_{c\in C}\mathbb{I}[\exists r\in R^\star:J(c,r)=1].$}
\end{equation}
\endgroup
Recall-side decisions are cached and reused only when they already establish
the relevant equivalence. Otherwise, the precision-side procedure invokes a
fresh judgment under its direction-specific prompt. We compute task-level
precision, recall, and $F_1$ before macro-averaging across tasks, so tasks with
larger reference rubrics do not dominate the aggregate. Because matching is
LLM-judged rather than independently human-validated, all reported coverage
numbers should be interpreted as conditional on this protocol.

\subsection{Greedy Reference-Informed Diagnostic}
\label{sec:greedy_diagnostic}

For each candidate $c$, define $M(c)=\{r\in R^\star:c\sim r\}$. Starting with
$S_0=\emptyset$, the diagnostic chooses
\begingroup\footnotesize
\begin{equation}
\label{eq:greedy_selection}
\adjustbox{max width=0.78\columnwidth}{$\displaystyle c_j^\star=\arg\max_{c\in P(t)\setminus S_{j-1}}\bigl|M(c)\setminus\bigcup_{\tilde c\in S_{j-1}}M(\tilde c)\bigr|.$}
\end{equation}
\endgroup
After each choice, set $S_j=S_{j-1}\cup\{c_j^\star\}$.
For $S\subseteq P(t)$, the set function
$f(S)=|\bigcup_{c\in S}M(c)|$ is monotone and submodular: selecting another
candidate cannot reduce coverage, and a candidate's marginal contribution
cannot increase as the selected set grows. Greedy maximization under a
cardinality budget therefore has the standard $(1-1/e)$ approximation
guarantee~\cite{nemhauser1978analysis}. Before the pool ceiling is reached,
greedy Recall@$k$ is a conservative estimate of optimal reference-informed
coverage, not an upper bound. At $k=60$, it reaches the measured 0.876
full-pool ceiling and is therefore optimal under the observed equivalence
labels. We use ``oracle'' only as shorthand for this non-deployable diagnostic;
it is not an alternative system and never changes the candidate pool.

\section{Experimental Setup}
\label{sec:setup_exp}

\paragraph{Benchmark and task selection.}
Our tasks are drawn from Harvey LAB, an open-source benchmark containing
more than 1,200 realistic legal-work tasks across multiple practice areas and
workflows~\cite{harveylab2026}. From this collection, we retain 20
document-grounded review and analysis tasks in legal and M\&A due diligence
for which the instruction, source-document data room, and hand-authored
reference rubric are available. We use 16 tasks for development and reserve
four for fully held-out evaluation. Held-out tasks contain 5--17 source
documents and 58--73 reference criteria. For example, one
held-out task asks the system to \emph{``Review the attached vendor materials,
draft MSA, and internal documents and prepare a due diligence memo for the
procurement committee.''} Its data room contains eight documents, including a
draft master services agreement, a business associate agreement, a SOC~2
summary, a procurement policy, internal evaluation emails, and vendor
materials. The corresponding rubric contains 73 criteria, including
\emph{``Executive Summary includes total contract value of \$43.2 million,''}
with a PASS/FAIL grading note. All held-out instructions, documents, criteria,
and numerical figures are excluded from development. We compare the full multi-axis pool with a GPT single-call baseline. The
baseline and our system use the same model and Azure deployment, comparing a single rubric-writing call against five distinct generation processes. Reference-free selection uses frozen \texttt{all-MiniLM-L6-v2} embeddings,
$K=5$-neighbor recurrence, and development-fixed bottom-30\% filtering. We
report macro recall, precision, $F_1$, and Recall@$k$. Complementarity uses only
tasks with complete shared outputs.

% \paragraph{Implementation and reproducibility.}
% All five generation processes, the GPT single-call baseline, and the
% equivalence judge use \texttt{gpt-5.6-sol} through the same Azure OpenAI
% deployment. Generation uses \texttt{seed=42} and JSON mode. The baseline calls
% the same generator model once on the same task instruction and source
% documents, holding provider, model, deployment, and random seed fixed while
% changing the inference procedure. Reference-free selection uses frozen
% \texttt{all-MiniLM-L6-v2} embeddings, $K=5$ neighbors, and bottom-30\%
% filtering; the free-form process unions four independent generation runs.
% Bootstrap intervals over criterion-level recall indicators are approximately
% $\pm0.02$--$0.04$. These controls are intended to isolate structural search
% and ranking from changes in model capacity or deployment.

\paragraph{Implementation and reproducibility.}
All five generation processes, the GPT single-call baseline, and the
equivalence judge use \texttt{gpt-5.6-sol} through the same Azure OpenAI
deployment. Generation uses \texttt{seed=42} and JSON mode. The baseline calls
the same generator model once on the same task instruction and source
documents, holding provider, model, deployment, and random seed fixed while
changing the inference procedure. Reference-free selection uses frozen
\texttt{all-MiniLM-L6-v2} embeddings, $K=5$ neighbors, and bottom-30\%
filtering; the free-form process unions four independent generation runs.
Bootstrap intervals over criterion-level recall indicators are approximately
$\pm0.02$--$0.04$.

\section{Results}
\label{sec:results}

\subsection{Coverage and the Coverage--Selection Gap}
\label{sec:main_results}
\label{sec:gap_results}

Table~\ref{tab:main_results} shows that multi-axis generation improves reference recall across every evaluated task family, increasing the equally weighted family average from 0.782 to 0.872.

\begin{table}[H]
\centering
\small
\caption{Reference recall across held-out due-diligence task families. Multi-axis generation improves coverage in every evaluated family.}
\vspace{7 pt}
\label{tab:main_results}
\begin{tabular}{lrrr}
\toprule
\textbf{Held-out task family} & \textbf{GPT} & \textbf{Ours} & $\mathbf{\Delta}$ \\
\midrule
Industry Analysis & 0.837 & \textbf{0.888} & \textbf{+.052} \\
Vendor Due Diligence & 0.849 & \textbf{0.863} & \textbf{+.014} \\
Governance Review & 0.661 & \textbf{0.864} & \textbf{+.203} \\
\midrule
Family average & 0.782 & \textbf{0.872} & \textbf{+.090} \\
\bottomrule
\end{tabular}
\end{table}

At $k=60$, greedy reference-informed ordering
reaches the measured full-pool recall ceiling
of 0.876, whereas reference-free ranking reaches
0.349, leaving a 52.7-percentage-point gap
(Table~\ref{tab:budget_gap}).

\begin{figure*}[H]
\centering
\begin{tikzpicture}
\begin{axis}[
  width=0.95\linewidth,
  height=0.35\textheight,
  xmin=7, xmax=410,
  ymin=0, ymax=1.02,
  xtick={10,20,30,60,100,200,390},
  xticklabels={10,20,30,60,100,200,All},
  ytick={0,.2,.4,.6,.8,1.0},
  xlabel={Selection budget $k$},
  ylabel={Reference recall},
  tick label style={font=\scriptsize},
  label style={font=\small},
  grid=major,
  grid style={draw=black!10, line width=.25pt},
  axis line style={draw=black!65, line width=.45pt},
  tick style={draw=black!65},
  legend style={
    font=\scriptsize,
    draw=none,
    fill=none,
    legend columns=2,
    at={(0.02,0.98)}, anchor=north west,
    /tikz/every even column/.append style={column sep=5pt}
  },
  clip=false,
]
\addplot[name path=oracle, very thick, color={rgb,255:red,32;green,86;blue,119}, mark=*, mark options={fill=white, line width=.8pt}, mark size=2.1pt]
  coordinates {(10,.445) (20,.662) (30,.816) (60,.876) (100,.876) (200,.876) (390,.876)};
\addlegendentry{Greedy reference-informed}
\addplot[name path=free, very thick, color={rgb,255:red,150;green,83;blue,39}, mark=square*, mark options={fill=white, line width=.8pt}, mark size=2.1pt]
  coordinates {(10,.077) (20,.144) (30,.243) (60,.349) (100,.531) (200,.722) (390,.876)};
\addlegendentry{Reference-free ranking}
\addplot[draw=none, fill={rgb,255:red,198;green,215;blue,225}, fill opacity=.64]
  fill between[of=oracle and free];
\addplot[dashed, color=black!60, line width=.65pt] coordinates {(8,.876) (404,.876)};
\node[anchor=south east, font=\scriptsize, text=black!58] at (axis cs:404,.876) {pool ceiling};
\draw[<->, >=stealth, line width=.7pt, color=black!70] (axis cs:60,.349) -- (axis cs:60,.876);
\node[fill=white, fill opacity=.92, text opacity=1, rounded corners=1pt, inner sep=2pt, font=\scriptsize\bfseries, align=center] at (axis cs:82,.61) {52.7 pp\\selection headroom};
\end{axis}
\end{tikzpicture}
\caption{Coverage--selection gap on the fixed held-out candidate pools.}
\label{fig:coverage_selection_gap}
\end{figure*}

\paragraph{Consistency across held-out tasks.}
The coverage--selection discrepancy is observed on every held-out task, not
only in the macro average (Table~\ref{tab:per_task_gap}). At $k=30$,
reference-free recall ranges from 0.203 to 0.293, compared with 0.781--0.864
for reference-informed greedy selection. At $k=60$, greedy attains each
task's measured full-pool ceiling (0.863--0.897), while the reference-free
ranker recovers only 0.315--0.373. These task-level differences make the
observed gap less dependent on the choice of aggregation.

\begin{table}[h]
\centering
\caption{Held-out task-level coverage. $|R^\star|$ denotes
reference-rubric size and $|P|$ candidate-pool size.
Greedy is reference-informed and diagnostic.}
\label{tab:per_task_gap}

\begin{tabular}{lrrrrrrr}
\toprule
& & &
\multicolumn{2}{c}{\textbf{Ref.-free}} &
\multicolumn{2}{c}{\textbf{Greedy}} &
\small
\textbf{Pool} \\
\cmidrule(lr){4-5}
\cmidrule(lr){6-7}
\textbf{Task} &
$\mathbf{|R^\star|}$ &
$\mathbf{|P|}$ &
\textbf{@30} &
\textbf{@60} &
\textbf{@30} &
\textbf{@60} &
\textbf{Ceil.} \\
\midrule
Industry 1 & 58 & 478 & .241 & .345 & .828 & .879 & .879 \\
Industry 2 & 58 & 384 & .293 & .362 & .793 & .897 & .897 \\
Vendor diligence & 73 & 362 & .233 & .315 & .781 & .863 & .863 \\
Governance review & 59 & 336 & .203 & .373 & .864 & .864 & .864 \\
\midrule
\textbf{Task macro} &
-- & -- &
\textbf{.243} &
\textbf{.349} &
\textbf{.816} &
\textbf{.876} &
\textbf{.876} \\
\bottomrule
\end{tabular}
\end{table}

% \begin{table}[h]
% \centering
% \footnotesize
% \caption{Held-out task-level coverage. $|R^\star|$ denotes
% reference-rubric size and $|P|$ candidate-pool size.
% Greedy is reference-informed and diagnostic.}
% \label{tab:per_task_gap}

% \setlength{\tabcolsep}{3pt}
% \renewcommand{\arraystretch}{1.1}

% \resizebox{0.88\columnwidth}{!}{%
% \begin{tabular}{lrrrrrrr}
% \toprule
% & & &
% \multicolumn{2}{c}{\textbf{Ref.-free}} &
% \multicolumn{2}{c}{\textbf{Greedy}} &
% \textbf{Pool} \\
% \cmidrule(lr){4-5}
% \cmidrule(lr){6-7}
% \textbf{Task} &
% $\mathbf{|R^\star|}$ &
% $\mathbf{|P|}$ &
% \textbf{@30} &
% \textbf{@60} &
% \textbf{@30} &
% \textbf{@60} &
% \textbf{Ceil.} \\
% \midrule
% Industry 1 & 58 & 478 & .241 & .345 & .828 & .879 & .879 \\
% Industry 2 & 58 & 384 & .293 & .362 & .793 & .897 & .897 \\
% Vendor diligence & 73 & 362 & .233 & .315 & .781 & .863 & .863 \\
% Governance review & 59 & 336 & .203 & .373 & .864 & .864 & .864 \\
% \midrule
% \textbf{Task macro} &
% -- & -- &
% \textbf{.243} &
% \textbf{.349} &
% \textbf{.816} &
% \textbf{.876} &
% \textbf{.876} \\
% \bottomrule
% \end{tabular}%
% }

% \end{table}
\paragraph{Aggregation of recall.}
Table~2 reports recall averaged equally across the
three task families, whereas Table~3 reports
macro recall averaged equally across the four
held-out tasks. Consequently, the family-level
average of 0.872 differs slightly from the
task-level macro recall of 0.876. Unless otherwise
specified, subsequent coverage--selection analyses
use task-level macro-averaged recall.

\paragraph{Sensitivity to the selection budget.}
The discrepancy persists beyond the two budgets highlighted above.
The reported gap is 36.8 percentage points at $k=10$, 51.8 points
at $k=20$, 34.5 points at $k=100$, and 15.4 points even at $k=200$
(Table~\ref{tab:budget_gap}). It narrows as the selected subset
approaches the full pool, consistent with coverage already present
but ranked too low by local semantic recurrence.

\begin{table}[h]
\centering
\small
\setlength{\tabcolsep}{4pt}
\caption{Held-out task-macro recall across selection budgets,
using the fixed candidate pools. Gaps are greedy minus
reference-free recall, in percentage points.}
\label{tab:budget_gap}
\begin{tabular}{rrrr}
\toprule
$\mathbf{k}$ & \textbf{Reference-free} & \textbf{Greedy} &
\textbf{Gap (pp)}\\
\midrule
10 & .077 & .445 & 36.8 \\
20 & .144 & .662 & 51.8 \\
30 & .243 & .816 & 57.3 \\
60 & .349 & .876 & 52.7 \\
100 & .531 & .876 & 34.5 \\
200 & .722 & .876 & 15.4 \\
\bottomrule
\end{tabular}
\end{table}

\subsection{Selection Remains the Bottleneck}
\label{sec:selector_results}

Table~\ref{tab:selection_ablations} compares the reference-free selection
ablations. Local semantic recurrence improves set-level F1 by 6.6 points, but
its candidate-level AUC is only 0.664. It measures pool-level recurrence, not cross-process convergence, and remains insufficient for tight-budget ranking. At $k=30$, greedy reference-informed ordering already reaches 0.816 recall, compared with 0.243 for the reference-free ranking, a 57.3-point gap. Across held-out tasks, greedy recall at this budget ranges from 0.781 to 0.864, while the corresponding full-pool ceilings range from 0.863 to 0.897.

\begin{table}[h]
\centering
\small
\caption{Selection ablations reported as change in set-level $F_1$. Positive values indicate improvement over the unfiltered candidate pool.}
\label{tab:selector_ablation}
\begin{tabular}{lr}
\toprule
\textbf{Selector} & $\mathbf{\Delta F_1}$ \\
\midrule
LLM item-wise pruning & -0.098 \\
LLM cluster representative & -0.098 \\
\textbf{Local redundancy ranking} & \textbf{+0.066} \\
\bottomrule
\label{tab:selection_ablations}
\end{tabular}
\end{table}

The two LLM-based alternatives remove useful tail criteria at nearly identical
rates, despite operating at different granularities. By contrast, local
redundancy supplies a modest aggregate improvement without solving the top-$k$
ordering problem. This distinction between filtering and ordering is important:
a selector can improve the quality of the retained set while still failing to
place complementary criteria early enough for a strict review budget.

\subsection{Structural Diversity, Not Volume}
\label{sec:complementarity_results}
\label{sec:axis_results}

Table~\ref{tab:axis_ablation} separates structural diversification from additional sampling: new structural axes add +0.060 to +0.100 recall, while volume expansions add at most +0.015.

\begin{table}[h]
\centering\small
\caption{Structural search axes yield substantially larger gains than additional generation volume.}
\label{tab:axis_ablation}
\begin{tabular}{lrr}
\toprule
\textbf{Modification} & $\mathbf{\Delta}$ \textbf{Recall} & $\mathbf{\Delta F_1}$ \\
\midrule
\multicolumn{3}{l}{\textit{New structural search axis}} \\
\quad Obligation axis & \textbf{+0.100} & \textbf{+0.050} \\
\quad Entity cross-reference axis & \textbf{+0.090} & \textbf{+0.050} \\
\quad Fact-to-schedule axis & \textbf{+0.060} & \textbf{+0.020} \\
\midrule
\multicolumn{3}{l}{\textit{Additional generation along existing directions}} \\
\quad Uncapped entity sweep & 0.000 & -0.050 \\
\quad Free-exploration computation & 0.000 & 0.000 \\
\quad Baseline-fact enumeration & +0.012 & -0.054 \\
\quad All volume expansions & +0.015 & -0.070 \\
\bottomrule
\end{tabular}
\end{table}

A development-task process diagnostic provides a complementary view
of the structural axes. On \texttt{shadow\_08}, the historically reported
reference-relevant yields were 40\% for obligation search, 39\%
for cross-document entity search, 25\% for free-form generation,
and 22\% for fact-to-schedule search. These yields overlap and
are not equivalent to the marginal ablation deltas in
Table~\ref{tab:axis_ablation}; the full development-task breakdown
is given in Table~\ref{tab:process_yields}. On the two tasks with complete shared outputs, GPT $\cup$ Ours reaches
\textbf{0.934 recall}, compared with 0.829 for GPT and 0.895 for Ours. The
union gain shows that even strong generators miss different requirements; it
also suggests that future work should emphasize mechanisms for identifying
novel marginal coverage rather than repeatedly sampling a single search view.

\section{Discussion}
\label{sec:discussion}
Rubric construction is a coverage-then-selection problem: multi-axis search first builds a broad pool, after which a selector must surface a compact and nonredundant subset. The 52.7-point gap at $k=60$ shows that these stages can have sharply different failure profiles. The generators have already located most reference requirements, but the deployable score does not recognize which candidates contribute new coverage. Reporting only full-pool recall would therefore overstate practical utility, while reporting only Recall@$k$ would incorrectly attribute ranking failures to generation. Structural axes outperform additional volume because they change what constitutes a unit of search: a contract provision, a cross-document entity, or a stated fact. Their gains are not reproduced by uncapping an existing sweep or adding more free exploration. 

\paragraph{Diagnostic error categories.}
A development-task element-strict analysis separates failures of evidence
specificity from failures of organization. On \texttt{shadow\_08}, the reported
residual misses included omitted named entities or numerical details (40\%)
and requirements associated with the wrong search axis despite being present
in the candidate pool (25\%). Other misses concerned external domain knowledge
(20\%) or whole-deliverable aggregate checks (15\%). These percentages come
from a single development task and a stricter judge, not the held-out test tasks
(Table~\ref{tab:failure_modes}).

\paragraph{Why local recurrence is insufficient.}
Our findings highlight a distinction between individual criterion relevance and the marginal coverage of a selected rubric. Conversely, a less frequently generated criterion may contribute substantially by capturing a requirement absent from the selected subset. The local recurrence score does not explicitly account for this interaction between selected criteria. Consequently, improving reference-free selection may require evaluating candidates jointly rather than ranking them solely through independent scores. Promising directions include evidence-aware facility-location objectives, learned pairwise novelty models, process-aware provenance retained through deduplication, and lightweight reviewer feedback. Such methods should be evaluated on the same fixed pools so that selection improvements are not confounded with additional generation.

\section{Conclusion}
\label{sec:conclusion}

We study document-grounded automatic rubric construction as a two-stage problem of candidate discovery and selection. Our multi-axis framework achieves high reference-criterion coverage by exploiting complementary structural views of source documents, with structural diversity proving more effective than additional generation volume. However, the substantial coverage--selection gap reveals that discovering relevant requirements does not guarantee their inclusion in a compact, usable rubric. Under limited selection budgets, the evaluated reference-free ranker fails to recover much of the coverage already present in the candidate pool, highlighting selection as a critical bottleneck. These findings motivate coverage-aware selection methods that prioritize complementary requirements over redundant criteria.

\section{Limitations}
\label{sec:limitations}

Our conclusions are bounded by four fully held-out M\&A tasks, despite the
larger development set. Criterion equivalence is judged by an LLM using the
same model family as the generators and has not been independently validated
by human annotators. The reference rubrics themselves encode practitioner
judgment rather than a unique ground truth, bootstrap intervals reflect
criterion-level rather than task-level variation, and cross-domain
generalization remains untested. The two-system union analysis is additionally
limited to the two tasks with complete shared outputs. These constraints make
the precise magnitudes provisional, but they do not erase the observed
diagnostic: within the evaluated document-grounded setting, a large share of
the missing top-budget coverage is already present in the pool.

\clearpage
\bibliographystyle{plainnat}
\bibliography{references}

\clearpage
\appendix

\section{Additional Dataset and Method Details}
\label{app:dataset_method}

\subsection{Held-out task inventory}
The four held-out test tasks are listed below to make the reported
task-level results traceable to the benchmark units.
These are the same four tasks evaluated in
Table~\ref{tab:per_task_gap}; they are not additional experiments.
Table~\ref{tab:heldout_ids} lists their identifiers and corresponding reference-rubric and candidate-pool sizes. The development-stage task inventory was described as eight drafting,
eight analysis, and four review tasks across the 20-task subset.
This is a within-domain task composition, not cross-domain evaluation.

\begin{table*}[h]
\centering
\small
\setlength{\tabcolsep}{4pt}
\caption{Documented held-out task identifiers and reference-rubric and
candidate-pool sizes.}
\label{tab:heldout_ids}
\begin{tabular}{p{0.55\linewidth}lrr}
\toprule
\textbf{Task identifier} & \textbf{Family} & $\mathbf{|R^\star|}$ & $\mathbf{|P|}$\\
\midrule
\texttt{draft-industry-summary\_\_sc01} & Industry & 58 & 478\\
\texttt{draft-industry-summary\_\_sc02} & Industry & 58 & 384\\
\texttt{draft-vendor-due-diligence-memorandum} & Vendor & 73 & 362\\
\texttt{review-governance-diligence\_\_sc02} & Governance & 59 & 336\\
\bottomrule
\end{tabular}
\end{table*}

\subsection{Requirement abstraction and metadata}
The development-stage requirement bank was described as 48 reusable
archetypes and 2,263 atomized rubric requirements. The atom schema
distinguishes a requested operation (\texttt{demands}: figure,
assessment, provision reference, presence, recommendation, date,
computation, benchmark, or party) from its applicable subject
(\texttt{applies\_to}: contracts, financial line items, deal,
deliverable, schedules, counterparties, named individuals, or properties).
The 119 computation templates are the computation-tagged slice of this
bank. These are historical method specifications; the full serialized
bank is not included in the available handoff materials. As an illustrative training example, a criterion about above-market
related-party rent can be decomposed into operations that extract the
contractual rate, identify a supported market benchmark, compare the
rates, and compute excess annual rent if the necessary quantities are
present. This illustrates abstraction rather than an independently
verified held-out recovery.

\section{Additional Development-Task Diagnostics}
\label{app:diagnostics}

\subsection{Generation-process yields}
\begin{table}[t]
\centering
\small
\caption{Historical process-level yields reported on the development
task \texttt{shadow\_08}. These yields overlap and are not marginal
leave-one-out ablation gains.}
\label{tab:process_yields}
\begin{tabular}{lr}
\toprule
\textbf{Generation process} & \textbf{Reported yield}\\
\midrule
Obligation search & 40\%\\
Entity cross-reference & 39\%\\
Free-form generation & 25\%\\
Fact-to-schedule search & 22\%\\
\bottomrule
\end{tabular}
\end{table}
The overlapping process yields in Table~\ref{tab:process_yields}
suggest that different search units produce reference-relevant
requirements. They are not additive coverage gains. Refinement
rewrites existing criteria and therefore is not assigned a
separate discovery yield.

\subsection{Tabular-evidence accessibility}
A historical document-rendering audit found that separately parsed
tables could be absent from the text seen by generation processes.
The reported average fraction of inaccessible source content was
6.3\%, reaching 19.1\% for the development task
\texttt{compare-target}. These percentages measure document visibility,
not rubric recall.

Separately, an element-strict evaluation on the development task
\texttt{shadow\_08} reported a recall increase from 0.684 to 0.763
after table rendering (a 7.9-percentage-point difference).
The evaluation criterion here is stricter than the requirement-level
equivalence protocol underlying the held-out results; these numbers
should not be pooled with the four-task test macro.

\subsection{Evidence specificity and reconciliation}
On \texttt{shadow\_08}, a historical specificity diagnostic
reported 0.90 figures per claim before refinement and 3.83 after
refinement. This indicates more numerical detail in generated
criteria, not independently validated correctness.

A methodological reconciliation example compares a narrative of
seven retention awards totaling \$1.85M with a supporting schedule
listing six awards totaling \$1.775M. A useful criterion would
require identifying the unreconciled difference instead of simply
restating either number. This example is illustrative, not a new
held-out case result.

\subsection{Residual failure analysis}
\begin{table}[t]
\centering
\small
\caption{Historical taxonomy of uncovered elements from
\texttt{shadow\_08}, assessed with an element-strict judge.
Percentages refer to residual misses on this development task,
not to the four held-out test tasks.}
\label{tab:failure_modes}
\begin{tabular}{p{0.67\linewidth}r}
\toprule
\textbf{Failure category} & \textbf{Share}\\
\midrule
Missing named entity or numerical detail & 40\%\\
External domain knowledge required & 20\%\\
Deliverable-level aggregate requirement & 15\%\\
In-pool requirement associated with wrong axis & 25\%\\
\bottomrule
\end{tabular}
\end{table}
The historical element-level analysis describes 30 remaining
misses, including 24 requiring numerical figures.
Table~\ref{tab:failure_modes} distinguishes evidence-specificity
failures from coverage and organization failures. Its categories
reflect one development task and a different judge protocol.

\section{Evaluation Implementation Notes}
\label{app:evaluation_details}

\subsection{Surviving recall-judge instruction}
The historical methodology preserves the following excerpt of the
recall-oriented matching instruction (not the full executable
system prompt):
\begin{quote}
\small
A candidate MATCHES a reference criterion when it states the
SAME REQUIREMENT: the same issue, about the same
entity/schedule/provision, calling for the same thing to be
identified, flagged, quantified or recommended.
\end{quote}
The reported recall-side procedure uses a nearest-candidate
shortlist and judges requirement equivalence rather than wording.

Precision uses a separately directed usefulness judgment, as
described in Section~\ref{sec:equivalence_protocol}; the
complete historical precision-side prompt was not recovered.
This excerpt should not be mistaken for full prompt reproducibility.

\subsection{Uncertainty estimation}
The historical protocol specifies 1,000 within-task bootstrap
draws over binary criterion-coverage indicators. The reported
aggregate uncertainty summarizes per-task interval half-widths;
it is not obtained by resampling four independent test tasks.
Consequently these intervals characterize criterion-level
uncertainty on the evaluated tasks, not task-level generalization
to new M\&A engagements. The original bootstrap samples and raw
judge-label matrices were not available for independent
recalculation.

\end{document}